\documentclass[onecolumn]{article}

\usepackage[preprint]{jmlr2e}

\jmlrheading{0}{2020}{0-0}{0/00}{00/00}{0000}{Alden Bradford, Tarun Yellamraju, and Mireille Boutin}

\ShortHeadings{Clustering small datasets in high-dimension
}{Bradford, Yellamraju, and Boutin}
\firstpageno{1}

\usepackage{url}
\usepackage{amsmath}
\usepackage{bm}
\usepackage{graphicx}
\usepackage{float}
\usepackage{amssymb}
\usepackage{subcaption}
\usepackage{tikz}   
\usetikzlibrary{positioning} 
\usepackage[capitalise, noabbrev]{cleveref} 
\usepackage{bbm} 

\begin{document}

\title{\Large Clustering small datasets in high-dimension by random projection 
\thanks{Funded in part by NSF grants EEC-1544244 and EEC-1826099.} }

\author{\name Alden Bradford \email bradfoa@purdue.edu \\
       \addr Department of Mathematics\\ Purdue University\\
       West Lafayette, IN 47907, USA
       \AND
       \name Tarun Yellamraju \email ytarun@purdue.edu \\
       \addr Department of Mathematics\\ Purdue University\\
       West Lafayette, IN 47907, USA
       \AND
       \name Mireille Boutin \email mboutin@purdue.edu \\
       \addr Department of Mathematics\\ Purdue University\\
       West Lafayette, IN 47907, USA}

\editor{}

\maketitle 

\begin{abstract}
Datasets in high-dimension do not typically form clusters in their original space; the issue is worse when the number of points in the dataset is small.
We propose a low-computation method to find statistically significant clustering structures in a small dataset. The method proceeds by projecting the data on a random line and seeking binary clusterings in the resulting one-dimensional data. Non-linear separations are obtained by extending the feature space using monomials of higher degrees in the original features. The statistical validity of the clustering structures obtained is tested in the projected one-dimensional space, thus bypassing the challenge of statistical validation in high-dimension. Projecting on a random line is an extreme dimension reduction technique that has previously been used successfully as part of a hierarchical clustering method for high-dimensional data. Our experiments show that with this simplified framework, statistically significant clustering structures can be found with as few as 100-200 points, depending on the dataset. The different structures uncovered are found to persist as more points are added to the dataset.

\end{abstract}


\section{Introduction}
\label{sect:introduction}

Clustering is a fundamental task in machine learning in which one seeks to 
organize sample points into similarity groups.
Different algorithms use different similarity criteria.
Since the choice of similarity criterion greatly affects the resulting groups, care must be taken to ensure that the criterion chosen for a given method is applicable to the data set under consideration.
Many clustering approaches work well for low-dimensional data, for example k-means \cite{kmeans}, Expectation Maximization \cite{dempster1977maximum}, BIRCH \cite{zhang1996birch} and DBSCAN \cite{ester1996density}. 
However clustering is considerably harder in higher dimensions. 
One reason for this is that small differences in information can be hidden under accumulated noise in several non-relevant dimensions. Another reason is the exponential growth of the space volume as the dimension increases. As a result, even a large number of points can be quite sparsely distributed in high dimension. Indeed, the volume of the inside of a unit cube in ${\mathbb R}^n$ approaches zero as $n$ approaches infinity, so the volume of the cube becomes concentrated along its surface. This implies that one can draw an infinite sequence of points inside an infinite dimensional unit cube, following any non-degenerated probability density function, with no accumulation point. In other words, a unit cube in ${\mathbb R}^\infty$ is not compact. To address this issue, high-dimensional data clustering methods (e.g., Proclus \cite{proclus}, Clique \cite{clique}, Doc \cite{doc}, Fires \cite{fires}, INSCY \cite{inscy}, Mineclus \cite{mineclus}, P3C \cite{p3c}, Schism \cite{schism}, Statpc \cite{statpc}, SubClus \cite{subclus}, and RP1D \cite{yellamraju2018clusterability} among others) seek to map the data to a lower-dimensional space, under various assumptions. As shown in \cite{boutin2019highly}, different mappings can yield distinct cluster groupings which represent the underlying structure equally well.

The goal of clustering is often not merely to partition a given set of points, but to be able to classify future data points as well. The data points are viewed as random samples drawn from some unknown distribution, and the sample points given are used to infer structures in this underlying distribution. If a criterion is found to separate the given samples into well-distinguished groups, it is important to check whether this grouping corresponds to an existing structure in the distribution or if it is just a random fluke. This can be done by computing the statistical significance of the result using an empirical statistical test. The fact that statistical tests are easier to perform in low-dimension is one more motivation for mapping the data to a low dimensional space.

\begin{figure}[ht]
\centering
\begin{tikzpicture}[scale=0.8]
\foreach \point in {
	(1.2,.8),
	(2.5,3.5),
	(2.8,6.4)
}{
	\filldraw \point circle [radius=0.15];
}

\foreach \point in {
	(3.4,0.2),
	(4.2,1.8),
	(4.8,3.6),
	(4.7,6.3),
	(5.8,4.8)
}{
	\draw \point circle [radius=0.15];
}

\draw[->,dashed,very thick] (1.6,3.3) -- (5.2,2.2)
	node[below,pos=0.4,sloped]{random vector};

\node at (3.5,7) {sparse data in high dimensions};
\node at (2.5,5.3) {cluster 1};
\node at (5,4) {cluster 2};

\draw[->,very thick] (7,3) -- (8.5,3);

\begin{scope}[shift = {(12,2.5)},scale=1.4]

\draw (-2,0) -- (2,0);
\draw[dashed] (0,-0.5) -- (0,2) node[pos=0,below]{threshold};

\draw (-2,0) cos (-1.5,0.7) sin (-1,1.4) cos (-0.5,0.7) sin (0,0);
\draw (0,0) cos (0.5,1) sin (1,2) cos (1.5,1) sin (2,0);

\node[below] at (-1,0) {cluster 1};
\node[below] at (1,0) {cluster 2};

\node[above] at (0,2) {histogram of projected data};

\end{scope}

\end{tikzpicture}
\caption{Sparse (high-dimensional) data can be divided into two groups after finding a projection on a line that yields a clear binary clustering.}
\label{fig:TARP_concept} 
\end{figure}
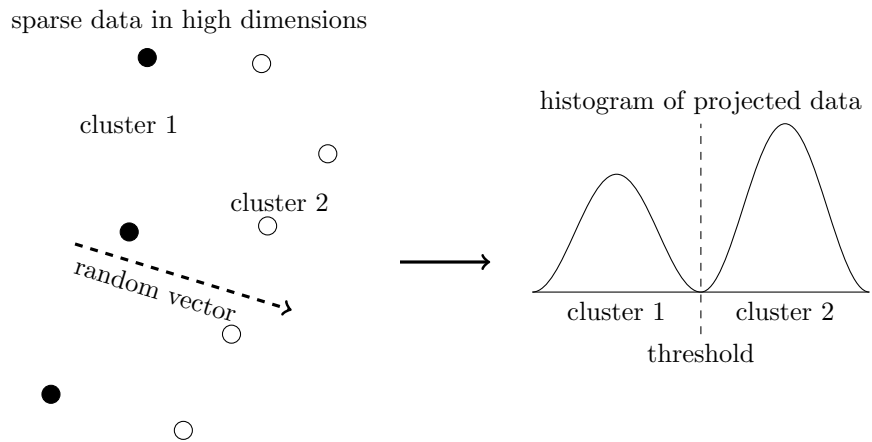

In this paper, we propose a random clustering method for high-dimensional data.
Our method is called $n$-TARP, where TARP stands for ``Thresholding After Random Projection," and is particularly well suited to cluster a small number of points in high-dimension (e.g., {\em fat} data). The core idea is to project the data onto random lines through the origin, and to set a threshold in the corresponding one-dimensional space so to separate the projected points into two groups. There is a grouping in the original high-dimensional space induced by this one-dimensional cluster assignment, though not necessarily a clustering, \cref{fig:TARP_concept}. Specifically, the threshold value and projection direction define a linear separation between the two groups in the original space. More generally, a non-linear separation can be obtained by extending the original feature space to a higher dimensional space with monomials in the original feature coordinates. 

This random projection and clustering is performed $n$ times, and the clustering that yields the best separation (in 1D) among those $n$ trials is picked. We subsequently test the statistical validity of the chosen clustering (in 1D). Neither of these steps is computationally expensive, and so this method scales up well with the space dimension and number of points. Given $m$ points in $d$ dimensions, we show in \cref{sect:method comments} that $n$-TARP takes $O(ndm\log m)$ time. Since the direction of projection is random, different clustering structures can be obtained in different runs. The statistical validity test insures that all the structures found are valid.

Other algorithms are specifically for large number of points. For example, consider SSC-Orthogonal Matching Pursuit (SSC-OMP). SSC-OMP uses the orthogonal matching pursuit algorithm for computing sparse representations. SSC-OMP can effectively handle 100,000 to 1,000,000 data points.


Below we discuss connections to existing work in Section \ref{sect:other work} before presenting our method in Section \ref{sect:method}. Numerical experiments follow in Section \ref{sect:experiments} before concluding in Section \ref{sect:conclusion}.

\section{Connections to Existing Work}
\label{sect:other work}
The proposed method is motivated by the observation that non-synthetic high-dimensional data often has a remarkably high likelihood of clustering when projected onto a random line \cite{han2015hidden}. We built on this in previous work to design a very simple hierarchical clustering method that surprisingly outperforms existing high-dimensional clustering algorithms when applied to real data \cite{yellamraju2018clusterability}. In this paper, we propose an even simpler clustering method with a view towards clustering small datasets in high-dimension.
As the small number of points limits the number of layers one can reliably obtain in a hierarchical clustering, our proposed method is not hierarchical. Since the issue of statistical validity is more crucial for small datasets, our proposed method includes a test for statistical validity.

Other previous work supports the use of 1D random projection to find relevant structures in a high-dimensional dataset. For example, Exploratory Projection Pursuit was proposed as a way to discover  ``non-linear" structures such as clusters in a dataset \cite{Friedman1987}. More generally,  random projections \cite{dasgupta1999learning, dasgupta2000experiments} have been proposed as a basis for 
dimensionality reduction techniques \cite{bingham2001random,three} with several applications in classification and clustering. For instance, \cite{two,five,six, seven} use random projections to reduce high-dimensional data into lower dimensional feature vectors for use with classifiers. Random projections have also been used in an iterative manner to find visual patterns of structure in data through dimension reduction \cite{four}. 

Many applications of random projections to dimensionality reduction \cite{bingham2001random} are motivated by the Johnson-Lindenstrauss lemma \cite{johnson1984extensions}, which states that a set of points in ${\mathbb R}^n$ can be projected to a lower dimensional space in such a way that the distances between the points are approximately preserved. For example, there are clustering methods based on random projections (e.g., \cite{fern2003random}) that project data to a lower dimensional space (but of dimensions greater than one) before assigning points to clusters based on their relative proximity.

As illustrated in Figure \ref{fig:TARP_concept},
our use of random projection has the opposite goal: by projecting down to a nearly trivial space, it seeks to dismantle the geometric structure defined by the pairwise distances. Rather than focusing on point proximity, it focuses on separation. Separation is sought in a one-dimensional space, where the points are most densely distributed, so that this separation can be reliably quantified. This distinguishes it from many other popular subspace clustering methods, for example SSC-OMP \cite{you2016scalable}, which relies on point proximity 
to group data points into clusters of various types.

\begin{figure}
\centering
	
	\begin{tikzpicture}[scale=0.5]
	\pgfmathsetmacro{\cubex}{4}
	\pgfmathsetmacro{\cubey}{4}
	\pgfmathsetmacro{\cubez}{4}
	\pgfmathsetmacro{\circsize}{0.75}
	
	\foreach \cornerpoint in {
		(0,0,0),
		(\cubex,0,0),
		(0,\cubey,0),
		(0,0,\cubez),
		(\cubex,\cubey,0),
		(\cubex,0,\cubez),
		(0,\cubey,\cubez),
		(\cubex,\cubey,\cubez)
	}{
		\filldraw \cornerpoint circle (5pt);
	}
	
	\draw (0,0,0) -- ++(\cubex,0,0) -- ++(0,\cubey,0) -- ++(-\cubex,0,0);
	\draw (0,0,0) -- ++(0,0,\cubez) -- ++(\cubex,0,0) -- ++(0,0,-\cubez);
	\draw (0,0,0) -- ++(0,\cubey,0) -- ++(0,0,\cubez) -- ++(0,-\cubey,0);
	\draw (\cubex,\cubey,\cubez) -- ++(0,0,-\cubez);
	\draw (\cubex,\cubey,\cubez) -- ++(0,-\cubey,0);
	\draw (\cubex,\cubey,\cubez) -- ++(-\cubex,0,0);
	
	\begin{scope}
	\pgftransformxslant{1}
	\pgftransformyshift{-2cm}
	\pgftransformyscale{-1}
	\pgfplothandlerlineto
	\pgfplotfunction{\t}{-1,-0.95,...,5}{\pgfpointxy{\t}{ 1*(exp(-8*(\t)^2)+exp(-8*(\t-\cubex)^2))}}
	\pgfusepath{stroke}
	\draw[dashed,thick] (-1,0) -- (5,0);
	\end{scope}
	
	\begin{scope}
	\pgftransformxshift{5cm}
	\pgftransformxslant{1}
	\pgftransformyscale{0.39}
	\pgftransformrotate{-90}
	\pgfplothandlerlineto
	\pgfplotfunction{\t}{-1,-0.95,...,5}{\pgfpointxy{\t}{ 2*(exp(-8*(\t)^2)+exp(-8*(\t-\cubex)^2))}}
	\pgfusepath{stroke}
	\draw[dashed,thick] (-1,0) -- (5,0);
	\end{scope}
	
	\begin{scope}
	\pgftransformyshift{-0.5cm}
	\pgftransformxshift{-2cm}
	\pgftransformrotate{90}
	\pgfplothandlerlineto
	\pgfplotfunction{\t}{-1,-0.95,...,5}{\pgfpointxy{\t}{ 2*(exp(-8*(\t)^2)+exp(-8*(\t-\cubex)^2))}}
	\pgfusepath{stroke}
	\draw[dashed,thick] (-1,0) -- (5,0);
	\end{scope}
	\end{tikzpicture}
\caption{A set of $2^p$ points in ${\mathbb R}^p$ are positioned on the corners of a unit cube. Projections along the axes of the cube reveal different possible clustering structures.}
\label{fig:many_projection_direction_cube}
\end{figure}
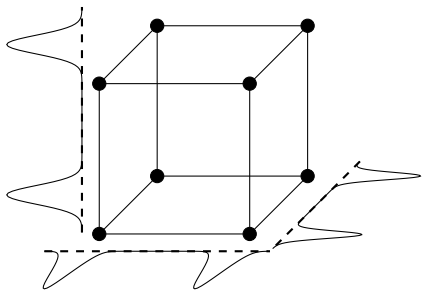

The most striking distinguishing characteristic of our approach is that it does not a priori assume that the data has a unique clustering structure. Instead, it assumes that there might be many different ways to cluster the data, and thereby attempts to generate several different clusterings. This may be slightly difficult to conceive, as the traditional notion of clusters consists of blobs (perhaps elongated or even forming complicated shapes) of densely distributed points with a clear separation in between. In this case, there is of course only one correct way to cluster the data. However, as pointed out in \cite{yellamraju2018clusterability}, such a structure appears to be incompatible with empirical observations. We show an alternative model in Figure \ref{fig:many_projection_direction_cube}. In this dataset, $2^p$ points are positioned on the corners of a unit cube in ${\mathbb R}^p$. Such a data set is very sparse and not clustered in any meaningful way. However, after projecting the points on any of the $p$ coordinate axes, a perfect binary clustering is revealed. This is the key idea behind our approach. We assume that the high-dimensional dataset may not have any cluster in the original space, but that clusterings can be revealed after projection onto (different) directions vectors. Each direction vector defines one direction of {\em information} and discards the other directions as {\em noise}. By choosing different information direction and discarding the rest, different clusterings are obtained. To give an analogy, imagine a classroom full of students. One might divide the class into two groups based on whether or not the students wear glasses, disregarding their other characteristics. Alternatively, one could also divide them based on whether they live on or off campus. There is no single correct way to divide the students; either is a correct way to define two distinguished groups.

\section{Method}
\label{sect:method}

    \subsection{Testing the clusterability of a dataset}
    \label{sect:clusterability}
    There are two questions which must be considered before n-TARP can be justified. The first question, discussed in \cref{sect:likely_clusters}, is whether the data are likely to exhibit clusters once projected. The second question, discussed in \cref{sect:meaningful_clusters}, is whether these projected clusters are representative of the true underlying data distribution, or are merely artifacts of our specific data sample.

\subsubsection{Are projected clusters likely?}
\label{sect:likely_clusters}

To see if clusters are likely to appear in projections of our data we follow the approach proposed in \cite{yellamraju2018clusterability}, summarized here. Given a collection of points $a_1,a_2,\dots,a_m\in \mathbb{R}$, and taking $a=\langle a_1,\dots,a_m\rangle^T$, we define ``normalized withinss" ($W$) as
\[
W=W(a)=\left(\frac{1}{\hat{\sigma}^{2} m}\right) \underset{C_{1},C_{2}}{\min} \left(
\sum_{a_i\in C_{1}}\left(a_{i}-\mu_{1}\right)^{2}+\sum_{a_i\in C_{2}}\left(a_{i}-\mu_{2}\right)^{2}\right),
\]
where $C_1$ and $C_2$ partition the set $\{a_1,\dots,a_m\}$ into groups with means $\mu_1$ and $\mu_2$ respectively, and $\hat{\sigma}^2$ is the empirical variance of $\{a_1,\dots,a_m\}$. Dividing by $\hat{\sigma}^2$ serves the purpose of obtaining scale invariance. The question of how to efficiently compute $W$ is considered in \cref{sect:compute_W}.

Given a data set packed into a matrix $X$, and a random projection vector $\mathbf{v}$, \cite{yellamraju2018clusterability} proposes studying the random variable $\mathbf{W}=W(X\mathbf{v})$. The idea is that low values of $W$ indicate a group of points which is well-split into two groups, so the data $X$ will be likely to be clustered by random projection if $\mathbf{W}$ takes on small values frequently.

\subsubsection{Are projected clusters meaningful?}
\label{sect:meaningful_clusters}

\begin{figure}
\centering
\includegraphics[width=0.6\textwidth]{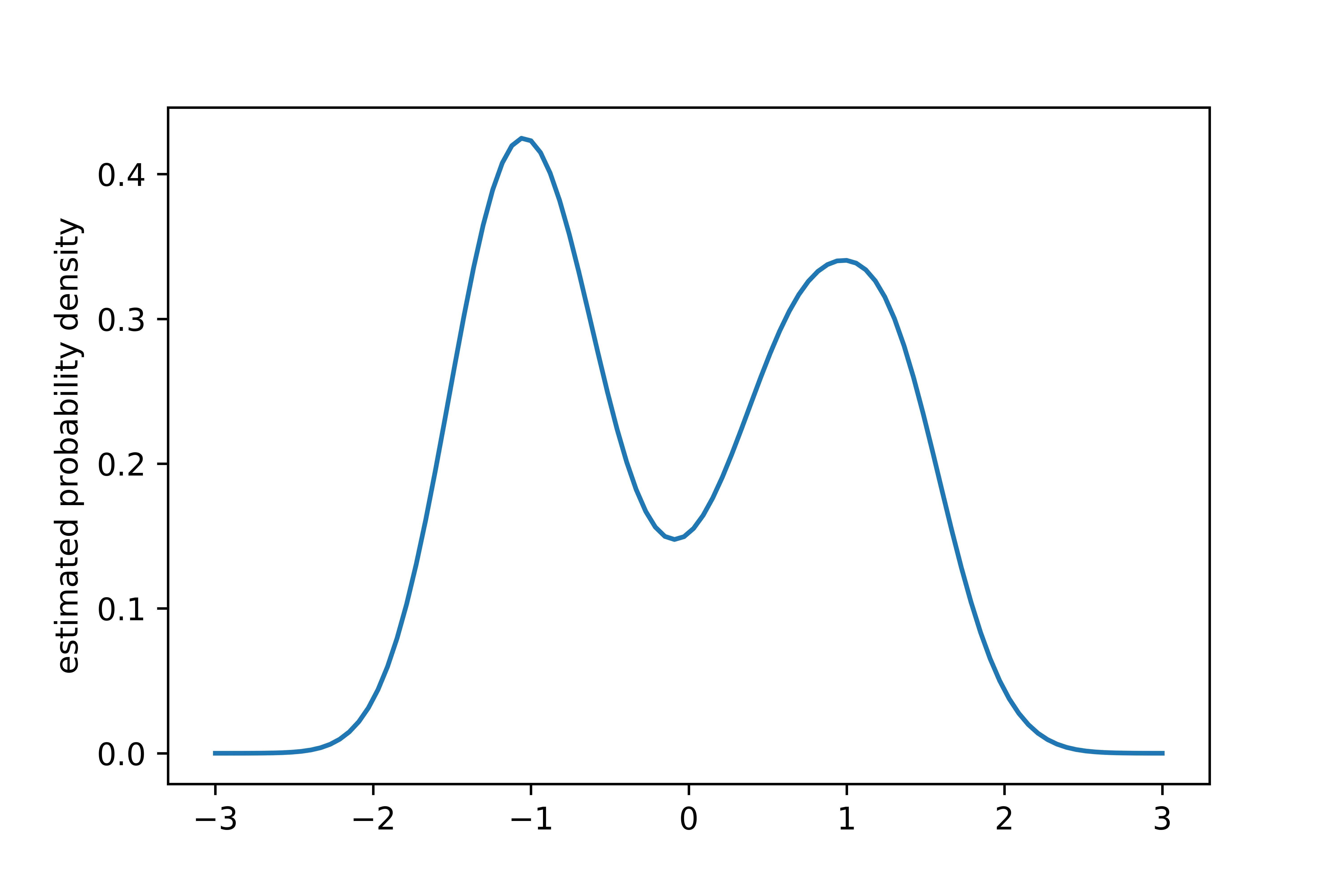}
\caption{A kernel density estimate for an unusual sample from a normal distribution. The pictured distribution has 50 points sampled from a standard Gaussian, and was chosen as the most-clustered among 1,000,000 such distributions.}
\label{fig:strange_normal}
\end{figure}

The fact that a (projection of a) sample set $X$ exhibits clusters does not necessarily imply that there are clusters in the distribution from which $X$ was drawn. As an example, consider \cref{fig:strange_normal}, which shows a kernel density estimate for a sample of 50 points drawn from a normal distribution. This particular sample has a normalized withinss of $W=0.147$. This was generated, for demonstration purposes, by simply drawing 1,000,000 such sample distributions and choosing the distribution with the smallest $W$. 
Although the empirical data is bimodal, the underlying normal distribution is not. Because of our small sample size, however, a bimodal structure was obtained ``by luck." This shows that it is not enough to simply find a projection with a small $W$, because the clustering generated by this separation does not necessarily generalize to other data drawn from the underlying distribution.

As was shown in \cite{yellamraju2018clusterability} the distribution of $\mathbf{W}$ depends heavily on the type of data considered. The question of finding the statistical significance of a given projection was discussed in \cite{SunJ.1991SLiE}, where they focus on classical projection pursuit. We address this issue by reserving some data for validation in a ``hold out" fashion; if a given projection cluster represents true structure in the underlying distribution of $X$, then a vector $v$ for which $W(Xv)$ is small should also cause $W(X'v)$ to be small, for another set of data $X'$ from the same source.

\subsubsection{Computing $W$}
\label{sect:compute_W}

\newcommand{\indi}{\ensuremath{\mathbbm{1}_{C_1}}} 

It will be helpful for computations later on to have a more thorough understanding of $W$. We can understand $W$ better by considering a random variable $Y$ which takes on the values $a_1,a_2,\dots,a_m$ with equal probability $\frac{1}{m}$. Then let $\indi$ denote a random variable indicating $Y\in C_1$. That is, $\indi=1$ if $Y\in C_1$ while $\indi=0$ otherwise. Now it is a direct consequence of the definition of conditional variance that 
\[
W = \left(\frac{1}{\text{Var}(Y)}\right)\underset{C_{1}}{\min} E[\text{Var}(Y|\indi)].
\]
Notice that the contribution of $C_2$ to $W$ is handled by the case where $\indi=0$. This allows us to understand $W$ in the context of the law of total variance which in this case states
\[
\text{Var}(Y)=E[\text{Var}(Y|\indi)]+\text{Var}(E[Y|\indi]).
\]
From this perspective, $W$ is precisely the fraction of the total variance in $Y$ that cannot be explained by a classifier of the form $\indi$. More familiar to the reader may be the complement of $W$, 
\[
\frac{\text{Var}(E[Y|\indi])}{\text{Var}(Y)},
\]
which is precisely the coefficient of determination $R^2$ used in regression analysis, in this case treating $Y$ as a function of $\indi$. That is,
\[
W=1-\underset{C_1,C_2}{\max} R^2.
\]
Besides connecting these two quantities, this also gives us a significantly easier way of computing $W$. Notice that $E[Y|\indi=1]=\mu_1$ while $E[Y|\indi=0]=\mu_2$. Hence,
\begin{align*}
    R^2
    &= \text{Var}(\mu_1 \indi + \mu_2 (1-\indi)),\\
    &= (\mu_1-\mu_2)^2\text{Var}(\indi),\\
    &= (\mu_1-\mu_2)^2 \tfrac{|C_1||C_2|}{m^2}.
\end{align*}
In dimensions 2 or greater, the problem of minimizing $W$ is NP-Hard; a good algorithm is presented in \cite{fast_k_means} in one dimension. Since we are interested only in the specific case of a binary clustering in one dimension and our number of data points $m$ is small, it is fast and practical to simply compute $R^2$ for every possible clustering $C_1,C_2$ and choose the best value. The clusters $C_1$ and $C_2$ must be convex, so in one dimension they must each lie within an interval; this means that the possible clusterings can be found by sorting the points $a_i$ and dividing by taking a threshold. In this way we can use an online method to compute all of the means, so computing all the $R^2$ values takes only $O(m)$ time. The most costly aspect of the procedure is sorting, which takes $O(m\log m)$ time.

\subsubsection{Typical values of $W$} \label{sect:typical_w_values}
In order to assess whether a given value of $W$ is significant we need to know what values would be typical given a sample from a typical distribution of points $X$. Here we provide a good approximation for the distribution of $\mathbf{W}$ assuming the points $X$ were drawn from a Gaussian distribution.

Notice that if $Z$ is a standard Gaussian random variable, then $E[\text{Var}(Z|Z<t)]$ is minimized at $t=0$. Furthermore, $\text{Var}(Z|Z<0)=\text{Var}(Z|Z\geq 0)=\text{Var}(|Z|)$. This provides an intuitive justification for approximating $W$ as
\[
W\approx \frac{\text{Var}(|Y|)}{\text{Var}(Y)}.
\]
We provide now a proof of the asymptotic behavior of this approximation for $W$. Again letting $Z$ be a standard Gaussian random variable, define 
\begin{align*}
\mu &= E[|Z|]=\sqrt{2/\pi},\\
\sigma^2 &= \text{Var}(|Z|)=1-2/\pi,\\
\kappa^2 &= E\left[\left((|Z|-\mu)^2-\sigma^2|Z|^2\right)^2\right]=8(\pi-3)/\pi^2.
\end{align*}
\begin{theorem}
Let $X_1,\dots,X_n$ be independent standard Gaussian random variables. Let $\overline{X}=\frac{1}{n}\sum X_i$ and let $\overline{X}'=\frac{1}{n}\sum |X_i|$. Then the quantity
\[
\frac{\sqrt{n}}{\kappa}\left[\frac{\frac{1}{n}\sum_{i=1}^n(|X_i|-\overline{X}')^2}{\frac{1}{n}\sum_{i=1}^n(X_i-\overline{X})^2} - \sigma^2\right]
\]
converges in distribution to a standard Gaussian random variable as $n\to\infty$.
\end{theorem}

\begin{proof}

Note that the law of large numbers guarantees that $\frac{1}{n}\sum_{i=1}^n(X_i-\overline{X})^2$ converges in distribution to the constant $\text{Var}(Z)=1$. Thus, we may apply Slutsky's theorem to simplify our task: we need only show the convergence of 
\[
\frac{\sqrt{n}}{\kappa}\left[\frac{1}{n}\sum_{i=1}^n(|X_i|-\overline{X}')^2 - \sigma^2\frac{1}{n}\sum_{i=1}^n(X_i-\overline{X})^2\right],
\]
and the result will follow. Now observe the following identities, which are easily confirmed:
\begin{align*}
\frac{1}{n}\sum_{i=1}^n(|X_i|-\overline{X}')^2 
&= \frac{1}{n}\sum_{i=1}^n(|X_i|-\mu)^2 - (\overline{X}'-\mu)^2,\\
\frac{1}{n}\sum_{i=1}^n(X_i-\overline{X})^2
&= \frac{1}{n}\sum_{i=1}^n X_i^2 -\overline{X}^2.
\end{align*}
These allow us to reduce our quantity further:
\[
\frac{\sqrt{n}}{\kappa} \sum_{i=1}^n \left[(|X_i|-\mu)^2-\sigma^2 X_i^2\right]-\frac{\sqrt{n}}{\kappa} \left[ (\overline{X}'-\mu)^2 - \sigma^2\overline{X}^2 \right]
\]
We claim the second term vanishes as $n\to \infty$. To see this, note $\sqrt{n}\overline{X}^2=(\sqrt{n}\overline{X})(\overline{X})$. By the law of large numbers, $\overline{X}$ converges in distribution to the constant 0. By the central limit theorem, $\sqrt{n}\overline{X}$ converges in distribution to a standard Gaussian. By Slutsky's theorem, their product must converge in distribution to the constant 0. The same procedure shows that $\sqrt{n}(\overline{X}'-\mu)^2$ vanishes as well.

Thus, all that remains is to show that 
\[
\frac{\sqrt{n}}{\kappa} \sum_{i=1}^n (|X_i|-\mu)^2-\sigma^2 X_i^2
\]
converges in distribution to the standard Gaussian. This is a direct application of the central limit theorem, noting that $E[(|Z|-\mu)^2-\sigma^2 Z^2] = 0$ and $\text{Var}\left((|Z|-\mu)^2-\sigma^2 Z^2\right)=\kappa^2$.

\end{proof}

\begin{figure}
\centering
\begin{subfigure}[b]{0.49\textwidth}
    \includegraphics[width=\textwidth]{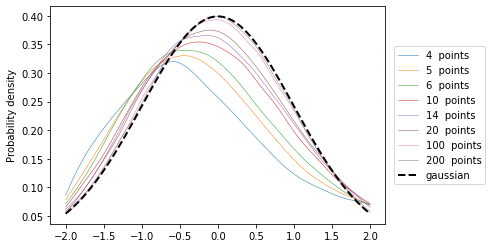}
    \caption{$\frac{\sqrt{n}}{\kappa}\left[\frac{\text{Var}(|Y|)}{\text{Var}(Y)} - \sigma^2\right]$}
    \label{fig:w_limiting_distribution_half_normal}
\end{subfigure}
\begin{subfigure}[b]{0.49\textwidth}
    \includegraphics[width=\textwidth]{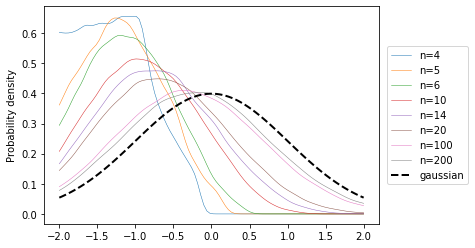}
    \caption{$\frac{\sqrt{n}}{\kappa}\left[\mathbf{W} - \sigma^2\right]$}
    \label{fig:w_limiting_distribution_unadjusted}
\end{subfigure}

\begin{subfigure}[b]{0.49\textwidth}
    \includegraphics[width=\textwidth]{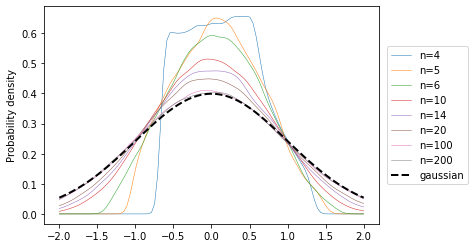}
    \caption{$\frac{\sqrt{n}}{\kappa}\left[\mathbf{W} - \sigma^2+\frac{1}{n}\right]$}
    \label{fig:w_limiting_distribution_centered}
\end{subfigure}
\begin{subfigure}[b]{0.49\textwidth}
    \includegraphics[width=\textwidth]{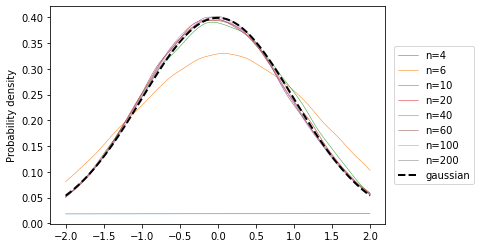}
    \caption{$\left(\mathbf{W} - \sigma^2+\frac{1}{n}\right)/\sqrt{\frac{\kappa^2}{n} - \frac{0.4}{n^{1.9}}}$}
    \label{fig:w_limiting_distribution_scaled}
\end{subfigure}
\caption{The empirical distributions of various functions of $W$ for various values of $n$ and 100,000 trials, using points drawn from a standard Gaussian distribution. This demonstrates that for $n>20$ the quantity $\left(\mathbf{W} - \sigma^2+\frac{1}{n}\right)/\sqrt{\frac{\kappa^2}{n} - \frac{0.4}{n^{1.9}}}$ is very nearly distributed as a standard Gaussian.}\label{fig:w_limiting_distribution}
\end{figure}

This allows us to approximate the distribution of $W$ accurately. To see this, compare the distributions shown in \cref{fig:w_limiting_distribution}. In this demonstration for each value of $n$ and 100,000 times, $n$ samples were drawn from a standard Gaussian distribution and the labeled quantity computed. We can see in \cref{fig:w_limiting_distribution_half_normal} and \cref{fig:w_limiting_distribution_unadjusted} that $\text{Var}(|Y|)/\text{Var}(Y)$ is in fact very close to $\mathbf{W}$ in terms of its limiting distribution. We can also see that while $\frac{\sqrt{n}}{\kappa}\left[\mathbf{W} - \sigma^2\right]$ does approach a standard Gaussian, it is biased for low values of $n$. We can correct for this bias by adding a factor which disappears as $n$ increases; this is shown in \cref{fig:w_limiting_distribution_centered}. We can go a bit further in correcting the variance for low $n$, accelerating this convergence with another small-$n$ correction demonstrated in \cref{fig:w_limiting_distribution_scaled}. This gives us a good rule for approximating $\mathbf{W}$ as a Gaussian random variable with mean $\sigma^2-\frac{1}{n}$ and variance $\frac{\kappa^2}{n}-\frac{0.4}{n^{1.9}}$.

Some justification is due for these empirically determined corrections. They were found in the following manner. First, several distributions of $\mathbf{W}$ were simulated, with 100,000 samples for each value of $n$ from 5 to 99. The mean and the variance were computed for each value of $n$. The deviation from the theoretical limiting mean of $\sigma^2$ and the theoretical limiting variance of $\kappa^2/n$ were recorded, and a linear regression was performed on a log-log scale. The corrections used are based on the best-fit lines found, with constants reported to one significant figure. The regressions for the mean and variance had $r^2$ values of $0.999$ and $0.998$ respectively.

This gives us a way to evaluate a $W$ score on other, possibly non-Gaussian data. For any value $w$ we can use this approximation to give a $p$ value for the probability $\mathbf{W}<w$ under the assumption the data were drawn from a Gaussian distribution.
    
    \subsection{The $n$-TARP method}
    \label{sect:method nTARP}
    The $n$-TARP method proceeds in two steps. Each step uses different data points, which are assumed to be independent. In the first step (the {\em observation} step) one performs $n$ binary clusterings by projecting onto a random line and thresholding. The best clustering among those $n$ is then selected as a potential clustering. The second step (the {\em validation} step), tests whether the projection direction of the potential clustering yields 1D sample points that are clustered in a statistically significant manner. Both steps use $W$ as a measure of the quality of a clustering in 1D.
Below we describe each step in detail. An implementation of this procedure can be downloaded from \cite{PURR2973}. 

Given are $m$ sample points $x_1,\ldots, x_{m} \in \mathbb{R}^d$. We  divide the points into two sets, with $m_1$ and $m_2$ points respectively, $m_1+m_2=m$. The first set $x_1,\ldots, x_{m_1}$ is used in the observation step. The second set, relabeled as $y_1,\ldots, y_{m_2}$, is used for the validation step.  

\begin{enumerate}
\item[Observation.] Only the points $x_1,\dots,x_{m_1}$ are used.
    \begin{enumerate}
    \item The following is done $n$ times, taking $i=1,2,\dots,n$:
        \begin{enumerate}
            \item A random direction vector $r_i\in \mathbb{R}^d$ is drawn from a Gaussian distribution.
            \item Each $x_j$ is projected onto $r_i$ by taking the dot product $z_{i,j} = x_j \cdot r_i$.
            \item The minimum withinss $W_i$ is computed for the set $\{ z_{i,1}, z_{i,2}, \dots, z_{i,m_1} \}$ according to the method described in \cref{sect:compute_W}
        \end{enumerate}
    \item The vector $r^*$ is chosen as the $r_i$ associated with the lowest computed $W_i$.
    \item The computation of the lowest $W_i$ naturally gives a partition of the $z_{i,j}$ into $C_1$ and $C_2$. Taking without loss of generality that $c_1<c_2$ for all $c_1\in C_1$ and $c_2\in C_2$, we define a threshold $t^*=\frac{1}{2}(\max(C_1)+\min(C_2))$.
    \end{enumerate}
\item[Validation.] The direction $r^*$ and threshold $t^*$ from the observation step, as well as the points $y_1,\dots,y_{m_2}$, are used to determine a $p$-score for the given clustering.
    \begin{enumerate}
    \item Project each $y_i$ onto $r^*$ by computing $y_i\cdot r^*$.
    \item Use the threshold $t^*$ to assign a cluster to each of the $y_i$ according to whether $y_i\cdot r^*<t^*$.
    \item Compute the withinss $W^*$ associated with that particular clustering; note this is not necessarily the same as the minimum withinss for the set $\left\{y_1\cdot r^*,\dots, y_{m_2}\cdot r^*\right\}$.
    \item Assign a $p$-value to this $W^*$ according to the approximation described in \cref{sect:typical_w_values}, based on the null hypothesis that the $y_i\cdot r^*$ are drawn from a Gaussian distribution.
    \end{enumerate}
\end{enumerate}
    
    \subsection{Comments and insights on $n$-TARP}
    \label{sect:method comments}

The complexity of the algorithm clearly scales linearly in the number of dimensions; similarly, the number of $n$-TARP trials $n$ is a linear factor. This, together with the discussion in \ref{sect:compute_W}, justifies the claim that $n$-TARP takes $O(ndm\log m)$ time to process $m$ points in dimension $d$.

The validation step in $n$-TARP is based on the null hypothesis that the projected validation data $y_i\cdot r^*$ are drawn from a Gaussian distribution. This is a reasonable assumption even when the original, unaltered data are not Gaussian. This is because of the theorem of \cite{DiaconisPersi1984AoGP} which states that, under some circumstances, most projections of a data set are nearly the same, and approximately Gaussian; since we are working with high-dimensional data we expect most projections to be nearly Gaussian.


Further, $n$-TARP can be viewed as a single basic unit similar to a single neuron/layer in a neural network. This unit can be combined or stacked together to make it more powerful. A tree structure based incorporation of $n$-TARP is presented in \cite{han2015hidden,yellamraju2018clusterability} while an extension of the same framework to the task of classification is presented in \cite{yellamraju2018benchmarks}. In this paper, we will stick with a single $n$-TARP unit, as our focus is on small data problems. 
Other architectures of clustering that combine several $n$-TARP units are appropriate for big data problems and the reader is referred to \cite{yellamraju2018clusterability} for some examples of the same. Those examples do not contain checks for statistical validity, rather, they consider clustering accuracy to measure effectiveness of the method. 

\section{Experiments}
\label{sect:experiments}
    
    \subsection{Data sets considered}
    \label{subsect:data_sets}
    Our main focus for $n$-TARP is on small data problems, wherein we have small number of data samples in a high-dimensional space. It is not immediately clear what types of structures we should expect to be able to uncover from such a dataset. With that in mind, the most reasonable test for the $n$-TARP method is how it performs on ``real-world'' data sets, i.e. clustering problems for which data is readily available. We will be using the following data sets. Unless otherwise specified, these were provided by the UCI Machine Learning Repository \cite{Dua:2019}.
    
    \begin{description}
    \item[m-feat:] the ``Multiple Features Data Set''. This consists of several different lists of features drawn from a set of 2000 handwritten numerals. For a detailed description of each, consult \cite{Dua:2019}. It is relevant that each sub-collection has a different number of dimensions for each data point, as summarized in this table:
    
    \begin{tabular}{r|llllll}
    mfeat- & fou & fac & kar & pix & zer &  mor \\\hline
    dimensionality & 76 & 216 & 64 & 240 & 47 & 6
    \end{tabular}
    
    Notice in particular that the mfeat-mor dataset has only six dimensions. In the experiments which follow it shows how the $n$-TARP method behaves differently when one of its key assumptions, i.e. that the data points lie in high dimensions, is violated.
    
    \item[libras:] the ``Libras Movement Data Set''. This consists of 360 points in 90 dimensions, representing measurements of hand position in recordings of Brazilian sign language.
    
    \item[mushrooms:] the ``Mushroom Data Set''. This consists of 8124 points, but is different from the other data sets in that it contains only categorical data. Each point is a list of 24 attributes measured from a mushroom. In order to embed the data into $\mathbb{R}^d$ a one-hot encoding is used, giving the resulting points a dimensionality of 119.
    
    \item[gaussian:] This is not a ``real'' dataset, rather it consists of 2000 points drawn from a standard Gaussian in 100 dimensions. This is provided as a null model, to show what happens in each experiment when given a collection with no structure.
    \end{description}



    \subsection{Experiments to determine how the parameter $n$ affects the successfulness of $n$-TARP.}\label{sec:experiment_n}
    
We wish to tell what value of $n$ makes a good choice. At one extreme, $n=1$ is not a good choice because it uses no information about the data set when selecting the direction; $1$-TARP is choosing the best direction from among only one direction. At the other extreme we would not want $n$ too large, because that increases linearly the time $n$-TARP takes.

We will measure the effect of $n$ on two important quantities. First, we investigate the frequency with which the $n$-TARP procedure is successful, that is, gives a low $p$-value in the validation step. Second, we will measure the extend to which successful (that is, validated with a low $p$ value) runs of $n$-TARP give clusterings which generalize to the rest of the data set. This is accomplished by performing the $n$-TARP validation step again, this time on a larger sample of the data set.

\subsubsection{The effect of $n$ on the $p$-values in the validation step}
Two-hundred points were selected at random and divided into a training set and a validation set, each of 100 points. The $n$-TARP procedure was followed 500 times and the resulting $p$-values stored. We report the fraction of those $p$-values which are less than 0.05, indicating a successful run.

\begin{figure}
\centering
\includegraphics[width=0.6\textwidth]{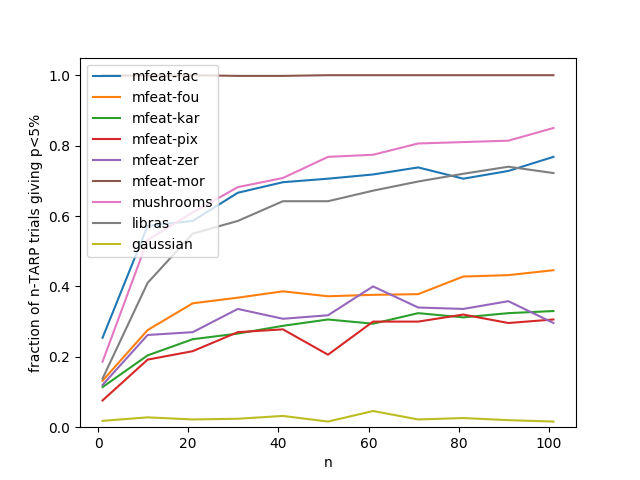}
\caption{How frequently significant clusterings are found for various values of $n$.}
\label{fig:expt_validation_vs_n}
\end{figure}

This procedure was carried out for each dataset and several values of $n$. The results are shown in \cref{fig:expt_validation_vs_n}. Observe that most of the benefit from increasing $n$ is found before around $n=60$. The benefit from increasing $n$ beyond this point is unclear.

\subsubsection{The effect of $n$ on the ability of clusterings to generalize to unseen data.} \label{sec:expt_generalization_vs_n}
As in the above experiment, two-hundred points are chosen at random and divided into a training set and a validation set each of 100 points, and the remaining points (however many there may be) are put into a testing set. $n$-TARP is performed repeatedly using the training and validation sets, until 500 successful (i.e. with a $p$ value less than 0.05) runs are completed. Note that this may mean that several more than 500 $n$-TARP trials are run in order to get these 500 successful trials. In order to ensure that this procedure terminates eventually, we never try more than 100 times to get a validated trial, instead aborting the experiment in this case. This did not occur here, but it will become relevant in \cref{sect:ExtFeat}.

For each trial, the given direction and threshold are used to cluster the testing data and a $p$-value is produced, as in the $n$-TARP validation step. The fraction of those $p$-values (computed from the testing set) which are less than 0.05 is reported.

\begin{figure}
\centering
\includegraphics[width=0.6\textwidth]{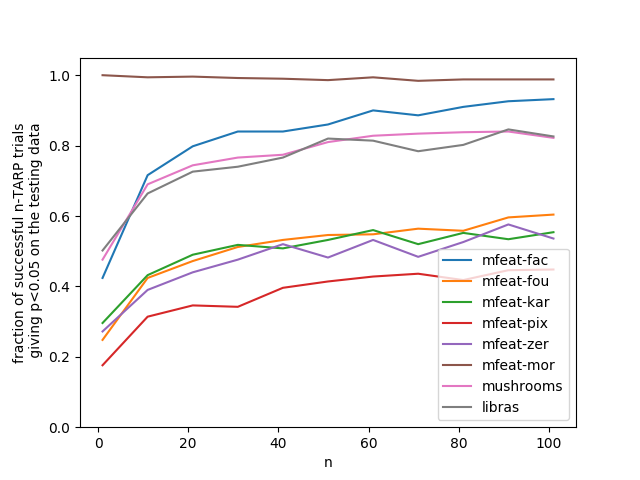}
\caption{How frequently generalizing clusterings are found for various values of $n$.}
\label{fig:expt_generalization_vs_n}
\end{figure}

This procedure was carried out for several values of $n$ and every data set except for the Gaussian data set, since that set did not produce enough successful trials. The results are shown in \cref{fig:expt_generalization_vs_n}. Notice that even for small $n$, the clusterings generalize to the unseen data most of the time. This implies the validation step is successful in removing any clusterings which do not generalize well. It is noteworthy that the Libras data set generalizes less well than the others. Still, with $n\geq 40$, nearly 90\% of the clusters obtained with $n$-TARP were found to persist when more data was added. Thus $n$-TARP is found to {\em scale up} very well, as nearly all clustering structures foudn in the small dataset exist in the larger set as well.

    \subsection{Experiments to determine how the number of data points affects the successfulness of $n$-TARP.}\label{sec:experiment_size}
    Here we repeat the same experiments as in \cref{sec:experiment_n}, but this time with $n$ fixed at 50 and varying the number of points in the training and validation data sets. For this experiment the dataset Libras was excluded because it did not have enough points for the range of sample sizes considered here.

\subsubsection{The effect of sample size on the $p$-values in the validation step}
2000 points were chosen at random from the dataset and assigned to be used for either training or validation, giving a pool of 1000 training and 1000 validation points. From each pool, 100 points were drawn initially to be used, and then 100 more points were added to the used data in each phase until the pools were exhausted.

\begin{figure}
\centering
\includegraphics[width=0.6\textwidth]{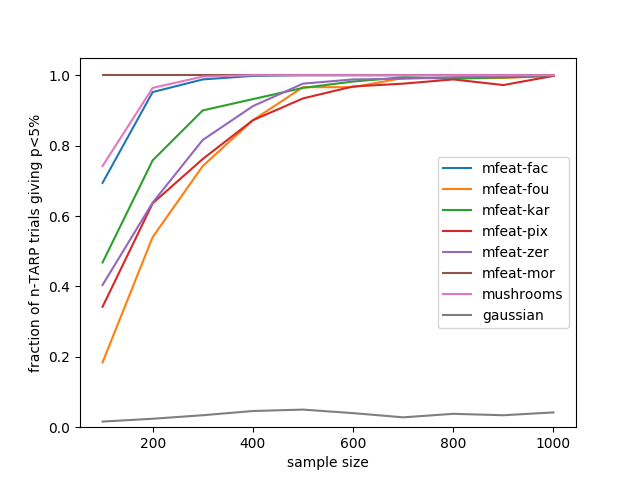}
\caption{How frequently significant clusterings are found for various sample sizes.}
\label{fig:expt_validation_vs_size}
\end{figure}

For each sample size, the $n$-TARP procedure was performed 500 times and the $p$-values recorded. We report the fraction of those $p$-values which are less than 0.05. The results are shown in \cref{fig:expt_validation_vs_size}. Notice that increasing the sample size uniformly improves the performance of $n$-TARP, though most of the improvement occurs by the time the sample size reaches 600 or so. After that point the benefit of increasing the sample size is significantly diminished. That being said, even for a sample size as low as 200 gives impressive performance, with all data sets giving statistically significant clusterings over 60\% of the time.

\subsubsection{The effect of sample size on the ability of clusterings to generalize to unseen data}
As in \cref{sec:experiment_n} the Gaussian dataset was excluded. 1000 points were drawn at random from the dataset to be used as a testing set. 400 more points were chosen and at random from the dataset and assigned to be used for either training or validation, giving a pool of 200 training and 200 testing points. From each pool, 20 points were drawn initially to be used, and then 20 more points were added to the used data in each phase until the pools were exhausted.

\begin{figure}
\centering
\includegraphics[width=0.6\textwidth]{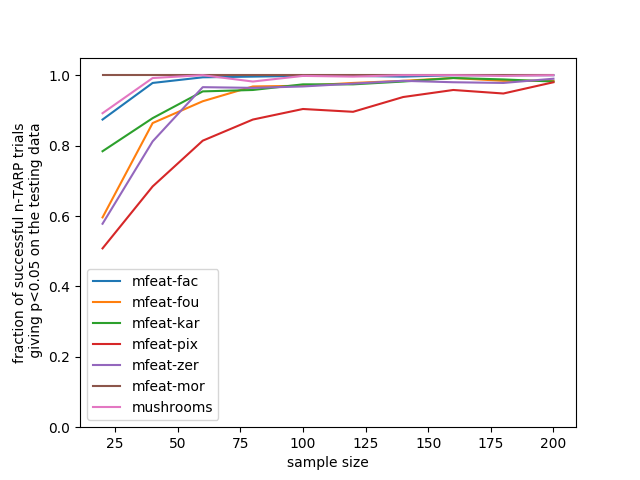}
\caption{How frequently generalizing clusterings are found for various sample sizes.}
\label{fig:expt_generalization}
\end{figure}

For each sample size, the $n$-TARP procedure was repeated until 500 successful trials were done, possibly requiring significantly more than 500 attempts. As in \cref{sec:expt_generalization_vs_n}, a maximum number of attempts was set after which the experiment would be aborted; in this case a maximum of 5000 attempts were made to get each successful trial. From those 500 successful trials the direction and threshold were used to cluster the 1000-point testing set, producing a $p$-value as in the $n$-TARP validation step. These 500 $p$-values were recorded and we report in \cref{fig:expt_generalization} the fraction of those $p$-values which are less than 0.05. As we can see from the graph, with as few as 100 points (50 observation, 50 validation), we get repeatable clustering in more than 90\% of cases.

Notice that we find that significantly fewer points are needed to produce clusterings which generalize well to the larger testing set. This mirrors what we saw in \cref{sec:expt_generalization_vs_n}, that the validation step does well at preventing poor clusterings from begin accepted.

    \subsection{Feature Space Extension}
    \label{sect:ExtFeat}
    One limitation of $n$-TARP as described thus far is that it can only identify linear separations of the data given. We can make the method more powerful, in the sense of making it possible to find more types of separations, by augmenting the data. Specifically, if one data point is of the form $(x_1, x_2,\dots, x_{p-1}, x_p)$ then we can get quadratic separations by replacing the data point in the method with one of the form $(x_1, x_2, \dots, x_{p-1}, x_p, x_1^2, x_1 x_2,\dots, x_p x_{p-1}, x_p^2)$. This gives us access to separations of the data by quadratic surfaces, in return for dramatically increasing the dimensionality of the data. Since $n$-TARP reduces the dimension of the data back to 1 by projection, this trade-off is acceptable. We can even extend the data to get higher order surfaces; we can get separations of order $r$ by extending the data into a space of dimension $\binom{p+r}{r}$. Note that this grows rapidly in $r$, and therefore it is not practical to use this method for even modest orders without careful memory management.

To see how extending the data in this manner affects the efficacy of $n$-TARP, we repeat each of the experiments from \cref{sec:experiment_n,sec:experiment_size}, this time first replacing each data point with its quadratic extension. 

\begin{figure}
\centering
\begin{subfigure}[b]{0.49\textwidth}
    \includegraphics[width=\textwidth]{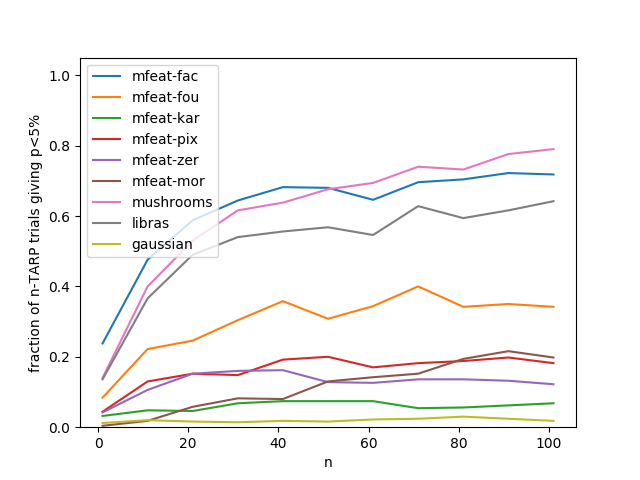}
    \caption{4.2.1}
    \label{fig:4_2_1_quadratic}
\end{subfigure}
\begin{subfigure}[b]{0.49\textwidth}
    \includegraphics[width=\textwidth]{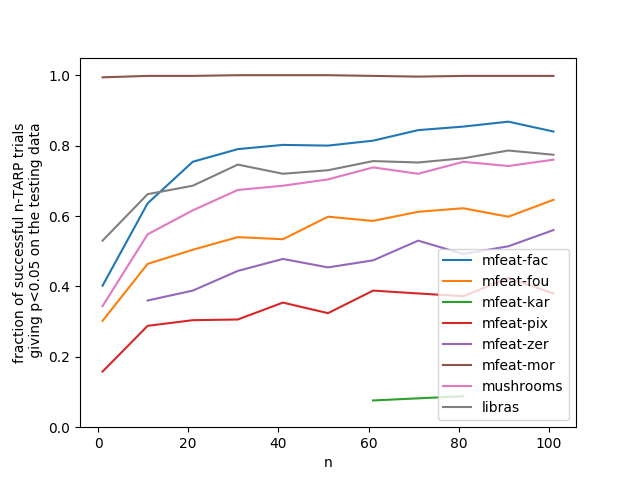}
    \caption{4.2.2}
    \label{fig:4_2_2_quadratic}
\end{subfigure}

\begin{subfigure}[b]{0.49\textwidth}
    \includegraphics[width=\textwidth]{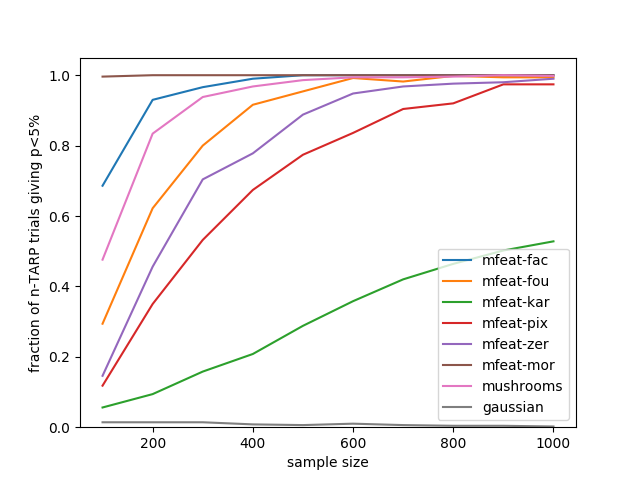}
    \caption{4.3.1}
    \label{fig:4_3_1_quadratic}
\end{subfigure}
\begin{subfigure}[b]{0.49\textwidth}
    \includegraphics[width=\textwidth]{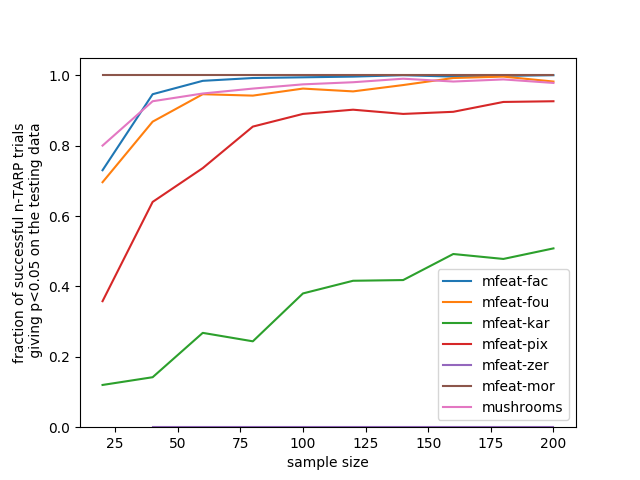}
    \caption{4.3.2}
    \label{fig:4_3_2_quadratic}
\end{subfigure}
\caption{The results from repeating each of the experiments in \cref{sec:experiment_n,sec:experiment_size}, this time with the data first extended quadratically.}\label{fig:quadratic_experiments}
\end{figure}

The results can be found in \cref{fig:quadratic_experiments}.
We can see that the performance is noticeably degraded across the board. Also, in the quadratic analogues of experiments 4.2.2 and 4.3.2, we some times were unable to get enough validated projections to continue the experiment; gaps in the plots reflect this. The general trends remain, however: an increase in $n$ or the number of points in training and validation data both lead to an increased number of validated separations, and an increase in the ability of those separations to generalize to unseen data. It is notable that one dataset in particular, mfeat-kar, did significantly worse when extended quadratically. This could be interpreted to mean that the structures to be found in mfeat-kar are linear in nature, and therefore are concealed rather than highlighted by exploring quadratic separation surfaces.

    \subsection{Comparison with other methods}
    Here we present an experiment which shows one way in which $n$-TARP is different from many other methods. Specifically, the built-in validation step is a reliable way of rejecting data sets which exhibit no clusters. Where other methods will find and report clusters, even in the absence of true underlying structures, $n$-TARP instead will indicate clearly that no reliable clusters are found by reporting a high $p$-value.
    
To demonstrate this effect, we use three types of synthetic data set. For uniformity, all three sets consist of 200 points in 100 dimensions. They are:
\begin{description}
\item[Gaussian:] Points are sampled independently from a standard Gaussian distribution.
\item[Uniform:] Each coordinate is sampled uniformly from $[0,1]$. This can be viewed as uniform sampling from a 100-dimensional hypercube. Then the dataset is rotated by multiplying by a random unitary matrix.
\item[Dilated cube:] Each coordinate $j$ (for $1\leq j \leq 100$) is a Bernoulli($\frac{1}{2}$) random variable, multiplied by a factor of $r^j$, for a fixed number $r$. We choose $r=1.1$. Then the dataset is rotated by multiplying by a random unitary matrix. For a further discussion of this model, see \cite{boutin2019highly}.
\end{description}
Note that each of these data models is random, so we can run many independent trials to observe typical behavior; this would otherwise pose a problem when trying to compare with deterministic algorithms, since by definition a deterministic algorithm would give us the same results for each trial if its inputs did not change.

For several popular clustering algorithms we measured their run time and how many clusters they found. We performed 100 trials with new data each time, and report the results in \cref{fig:run_times} and \cref{fig:cluster_counts}.

First consider \cref{fig:run_times}. This shows that all algorithms considered took a comparable amount of time to run, though the hierarchical, dbscan, and n-TARP reliably performed the fastest by a decent margin.

Now consider \cref{fig:cluster_counts}. This demonstrates how most of the algorithms available will always report however many clusters they are asked for. The exceptions to this are affinity propogation, optics, and n-TARP. Affinity propogation performs poorly here, consistenly reporting many clusters for the Gaussian and Uniform data models, where there should only be one cluster reported. This makes sense, because affinity propogation relies on computing pairwise distances which are unreliable in high dimensions. Optics performs very well; it correctly identifies only one cluster in the gaussian and uniform cases, while identifying lots of structures in the dilated cube model. This also is reasonable; optics is hard-coded to look for structures aligned with the axes, and all the structure in the dilated cube model is aligned with its axes. n-TARP performed acceptably well on this test; it assigned only one cluster most of the time to the gaussian and uniform data models, and identified more than one cluster in the dilated cube model. It is worth noting that this version of n-TARP has only two possible outcomes: either the validation step was successful and two clusters are identified, or the validation step was unsuccessful and the entire data set is assumed to belong to the same cluster. In principle n-TARP could then be chained to identify sub-clusters, but that was not part of this experiment.

\begin{figure}
\begin{tabular}{llll}
\hline
{} &     gaussian &      uniform &         cube \\
\hline
kmeans-2             &  0.20 (0.06) &  0.20 (0.05) &  0.15 (0.04) \\
kmeans-3             &  0.25 (0.07) &  0.25 (0.07) &  0.20 (0.05) \\
kmeans-5             &  0.33 (0.08) &  0.31 (0.09) &  0.29 (0.08) \\
kmeans-10            &  0.46 (0.11) &  0.45 (0.10) &  0.47 (0.11) \\
affinity\_propagation &  0.33 (0.21) &  0.42 (0.26) &  0.39 (0.21) \\
mean\_shift           &  1.15 (0.16) &  1.10 (0.17) &  1.32 (0.15) \\
spectral\_clustering  &  1.86 (3.76) &  0.43 (0.08) &  0.41 (0.08) \\
hierarchical         &  0.04 (0.02) &  0.06 (0.00) &  0.06 (0.01) \\
dbscan               &  0.05 (0.04) &  0.10 (0.04) &  0.07 (0.03) \\
optics               &  0.59 (0.15) &  0.67 (0.10) &  0.71 (0.08) \\
birch                &  0.13 (0.04) &  0.13 (0.04) &  0.14 (0.04) \\
n\_tarp               &  0.06 (0.01) &  0.06 (0.00) &  0.06 (0.01) \\
\hline
\end{tabular}
\caption{The mean and standard deviation among the 100 run-times (in seconds) for each algorithm on each synthetic dataset.}
\label{fig:run_times}
\end{figure}

\begin{figure}
\begin{tabular}{llll}
\hline
{} &         gaussian &          uniform &             cube \\
\hline
kmeans-2             &        2 - 2 - 2 &        2 - 2 - 2 &        2 - 2 - 2 \\
kmeans-3             &        3 - 3 - 3 &        3 - 3 - 3 &        3 - 3 - 3 \\
kmeans-5             &        5 - 5 - 5 &        5 - 5 - 5 &        5 - 5 - 5 \\
kmeans-10            &     10 - 10 - 10 &     10 - 10 - 10 &     10 - 10 - 10 \\
affinity\_propagation &    10 - 14 - 200 &    14 - 20 - 200 &    24 - 27 - 200 \\
mean\_shift           &        1 - 1 - 1 &        1 - 1 - 1 &        1 - 1 - 1 \\
spectral\_clustering  &        8 - 8 - 8 &        8 - 8 - 8 &        8 - 8 - 8 \\
hierarchical         &        2 - 2 - 2 &        2 - 2 - 2 &        2 - 2 - 2 \\
dbscan               &  200 - 200 - 200 &  200 - 200 - 200 &  200 - 200 - 200 \\
optics               &        1 - 1 - 1 &        1 - 1 - 1 &    1 - 178 - 196 \\
birch                &        3 - 3 - 3 &        3 - 3 - 3 &        3 - 3 - 3 \\
n\_tarp               &        1 - 1 - 2 &        1 - 1 - 2 &        1 - 2 - 2 \\
\hline
\end{tabular}
\caption{The minimum, median, and maximum cluster count among the 100 trials for each algorithm on each synthetic dataset. In each case, the correct number of clusters is one.}
\label{fig:cluster_counts}
\end{figure}

\section{Conclusions}
\label{sect:conclusion}

In this paper, we introduced a novel clustering method called $n$-TARP, which is non-deterministic and based on point separations instead of point proximity. The method is designed to find multiple statistically significant binary clusterings rather than a unique ``best" grouping of data into clusters. As explained in \cite{boutin2019highly}, such a structure is more appropriate for real data. 
Our experiments show that it will find statistically significant clusters with as few as 200 data points in high-dimension. The method has a very low computational cost and can also be used for big data problems as well as low-dimensional problems.

The central idea of the method is the projection of data onto randomly generated vectors. The clustering is performed on the data projected on the line spanned by this vector, thus reducing the task to a one-dimensional clustering problem.
Our previous work \cite{han2015hidden,yellamraju2018clusterability} has shown that data in high-dimensional space has hidden structure that can be uncovered through this projection process. The experiments performed in this performed in this paper reconfirm this. 

This method includes a statistical validity evaluation (in the projected 1D space), which insures that the structures found persist when more data is added to the dataset (i.e., the clusters {\em scale up}). An extension step in which monomials in the feature coordinates are concatenated to the original feature vectors allows us to find non-linear separations as well.



Our work shows that even small datasets in high-dimensions have structures that can be found by projection on a random line. In accordance with the cube model of \cite{boutin2019highly}, this structure manifests itself as point separations in random projected 1D subspaces, leading to several distinct binary clustering. The cluster assignments themselves should be viewed as random variables induced by the random projections of $n$-TARP. More work is needed to fully understand these cluster distributions.


\bibliography{refs}

\end{document}